# A Flexible Online Classifier using Supervised Generative Reconstruction During Recognition.

Tsvi Achler [1]
E-mail: achler@gmail.com

**Abstract**

Matching the brain's flexibility and ability to quickly incorporate new information remains difficult using recognition algorithms. In addition, recognition scenarios in which humans require more time to recognize do not inherently require more time in recognition algorithms. This work proposes a recognition algorithm that uses dynamics in an opposite way from traditional recognition algorithms. Recognition algorithms also know as classifier algorithms function in two phases: a learning phase and a testing phase. Most classifier algorithms utilize feedforward connections during testing. They use dynamic feedforward-feedback signals to learn the feedforward connections during the learning phase.

A supervised generative classifier is proposed where the key innovation is that the dynamic feedforward-feedback component is essential during the testing as opposed to the learning phase. This changes the form which learned information is stored. The new form represents the fixed points or solutions of the network. This form allows for more intuitive symbolic-like weighs, more-flexible online learning, and the dynamics involved in the recognition phase emulates cognitive phenomena. Brain-like architecture, capabilities, and limits associated with this model suggest the brain may perform recognition and store information using a similar approach.

## Introduction

Recognition is an essential foundation upon which cognition and intelligence is based. Without recognition the brain cannot interact with the world e.g.: form internal understanding, memory, logic, display creativity, or reason. Subsequently the form of recognition information, as determined through the underlying neuron architectures and connection weights, can greatly affect memory and cognition. For example, whether sound is processed in the frequency or time domain will affect the ease of storing, comparing, and recognizing certain aspects of sound. Analogously certain forms of connections and weights may be more optimal for flexible learning.

Many recognition algorithms have been developed over the past half century. For the most part, they share a commonality that 1) they perform classification based on a simple use of feedforward weights during testing and 2) the feedforward weights are determined during learning though dynamic gradient descent methods (using feedforward-feedback methods). Recognition methods can be labeled based on their testing configuration. These classifiers are designated as feedforward.

Feedforward methods are known to have difficulties with online learning: updating or changing information independently. Such changes can lead to *Catastrophic interference* and *catastrophic forgetting* e.g. (French 1999; McClelland, McNaughton, & O'Reilly 1995). Without retraining old information, information can be quickly lost. Moreover the strategy of batch learning, learning all possible patterns at once, is not scalable for large data sets (Bottou & Le Cun 2004). Although some strategies have been proposed to ameliorate this problem, it is avoided here by modifying the form of information.

The key innovation presented here is a supervised recognition algorithm that uses a feedforward-feedback configuration to implement dynamics that converge through gradient descent *during testing*. This allows learning to be much simpler using symbolic-like expectations and the weights to represent fixed points. Moreover, changes required for online learning are localized when information is stored using fixed points, thus online problems are avoided.

**Table 1:** Comparisons between feedforward and feedforward-feedback supervised classifiers

| Structure During Recognition | Dynamics During | Weights | Relations | Modification of Fixed-Points | Inherent Cognitive Phenomena |
|---|---|---|---|---|---|
| Feedforward | Learning | **W** | Sub-Symbolic | Distributed | No |
| Feedforward-Feedback | Testing | **M** | Symbolic | Localized | Yes |

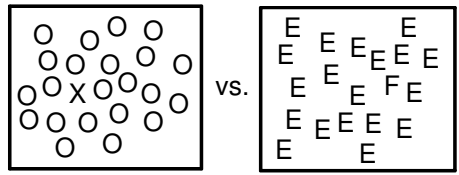

**Figure 1, Properties of Human Recognition.** It is faster to find the single pattern in the left panel than in the right panel. The feedforward-feedback classifier inherently displays analogous phenomena due to dynamics.

**Background**

Although many recognition algorithms have been developed over the past half century, supervised recognition systems share a commonality that they predominantly utilize a feedforward architecture during testing. Based on feedforward weights **W** they solve the recognition relationship:

$$\mathbf{Y} = \mathbf{WX} \quad \text{or} \quad \mathbf{Y} = f(\mathbf{W}, \mathbf{X}) \qquad (1)$$

Vector **Y** represents the activity of a set of labeled nodes that may be called output neurons, or classes in different literatures and individually written as $\mathbf{Y} = (Y_1, Y_2, Y_3, \ldots Y_H)^T$. They are considered supervised because the nodes can be labeled for example: $Y_1$ represents *"dog"*, $Y_2$ represents *"cat"*, and so on. Vector **X** represents sensory nodes that sample the environment, or input space to be recognized, and are composed of individual features $\mathbf{X} = (X_1, X_2, X_3, \ldots X_N)^T$. The input features can be sensors that detect edges, lines, frequencies, kernel features, and so on. **W** represents a matrix of weights or parameters that associates inputs and outputs. The relationship **WX** calculates the output using the feedforward weights and inputs. Thus the direction of information flow *during recognition* is feedforward: one-way from inputs to the outputs.

Learning **W** weights requires feedback in order to compare outputs to supervised training instances and project error signals back using methods such as the delta rule. However once **W** is defined, recognition during testing is feedforward. Variations on this theme can be found within different algorithm optimizations, for example: single-layer Perceptrons (Rosenblatt 1958), multilayer Neural Networks with nonlinearities introduced into calculation of **Y** (Rumelhart & McClelland 1986), and machine learning methods such as Support Vector Machines (SVM) with nonlinearities introduced into the inputs through the "kernel trick" (Vapnik 1995). Although these algorithms vary in specifics such as nonlinearities determining the function *f*, they share the commonality in that recognition involves a feedforward transformation using **W** during recognition.

Some feedforward algorithms include lateral connections for competition between output nodes **Y**. One variant, Adaptive Resonance Theory (Carpenter & Grossberg 1987) measures a goodness-of-fit after a winner-take-all competition. However such competition methods still rely on initially calculating **Y** node activities based on feedforward-trained weights **W** and to not iteratively modify the input layer activity as part of classification. Thus they fall into the feedforward category.

Recurrent networks are feedforward networks in a hierarchy where a limited number of outputs are also used as inputs. These networks can be unfolded into a recursive feedforward network e.g. (Schmidhuber 1992; Williams & Zipser 1994; Boden 2006). Thus they also fall into a feedforward category.

Networks based on feedforward-feedback structures are auto-associative networks and generative models. Generally, a Lyapanov function can be written and the networks converge to stable fixed-points. Typically in auto-associative networks, the feedforward and feedback connections have the same weights; however there are exceptions (McFadden et al 1992; Bogacz & Giraud-Carrier 1998). In generative models the feedback weight is the opposite sign of the feedforward weight. Historically auto-associative networks have been used during testing while supervised generative models are used to aid learning.

Two famous auto-associative networks are Anderson's "brain-state-in-a-box" and Hopfield networks (Anderson et al 1977; Hopfield 1982). In these networks when part of a learned input pattern is given, the network can complete the whole pattern through a dynamic process. However, the auto-associative networks do not perform supervised classification and solve equation 1 above.

Generative models are used to generate patterns and subtract them from the inputs. This is a strategy used to optimize learning of Bolzman (binary activation) networks e.g. (Hinton & Salakhutdinov 2006) and unsupervised networks where outputs are unlabeled e.g. (Olshausen & Field 1996). However these methods do not use the feedforward-feedback component during testing of supervised networks. For example it common to see an unsupervised generative network followed by a supervised feedforward classifier e.g. (Zieler et al 2010).

The recognition method proposed here functions like an auto associative network - during recognition - but uses a generative type network. It solves the supervised recognition problem represented by equation 1 but its weights are symbolic-like fixed-point solutions. Let's derive how it works.

**Mathematical Formulation**

Equation 2 of a linear perceptron can be rewritten using an inverse of equation 1:
$$W^{-1}\vec{Y} = \vec{X} \tag{2}$$
Lets define matrix **M** as the inverse or pseudoinverse of matrix **W**. The relation becomes:
$$M\vec{Y} - \vec{X} = 0 \tag{3}$$
Models founded on this equation can be referred to as generative models because they "generate" a reconstruction. The term **MY** is an internal prototype of input pattern(s) constructed using learned information **M**. When this reconstruction matches patterns present in **X** the equation goes to zero and recognition is accomplished. Information flows from **Y** to **X** using **M** and recognition is determined in the input or **X** domain. This flow of information describes top-down feedback, the opposite direction of feedforward. Equations 3 and 1 are the same, so the same **Y**'s should match both the feedforward and feedback equations and should both have the same fixed-points or solutions. This duality suggests that analogous network connectivity can be described in both feedforward and feedback manners.

However equation 3 does not provide a way to project input information to the outputs. To get around this, we use dynamic networks or equations that converge to equation 3.

One method that can be used is based on Least Squares to minimize the energy function: $E=\|X - MY\|^2$. Taking the derivative relative to **Y** and solving the equation becomes:
$$\frac{d\vec{Y}}{dt} = M^T\left(M\vec{Y} - \vec{X}\right) \tag{4}$$
This equation can be iterated until steady state, d**Y**/dt=0 (with no weights adjusted) resulting in the fixed point solution that is equivalent to **Y=WX**. Both feedforward and feedback connections are designated by **M**. This method is described as feedforward-feedback because **MY** transforms **Y** information into the **X** domain, thus a feedback process. **M**$^T$ transforms **X** information into the **Y** domain, thus a feedforward process.

Another way to converge to equation 3 is to use Regulatory Feedback (RF) e.g. (Achler 2012; Achler & Bettencourt 2011). The equation can be written as:

$$\frac{d\vec{Y}}{dt} = \vec{Y}\left(\frac{1}{V}M^T\left(\frac{\vec{X}}{M\vec{Y}}\right) - 1\right) \quad \text{where} \quad V = \sum_{j=1}^{N} M_{ji} \quad (5)$$

Using alternative notation, this can be written as:

$$\frac{dY_i}{dt} = \frac{Y_i}{\left(\sum_{j=1}^{N} M_{ji}\right)} \sum_{k=1}^{N} M_{ki} \left(\frac{X_k}{\sum_{h=1}^{H} M_{kh}Y_h}\right) - Y_i \quad (6)$$

where $M_{NxH}$ are the dimensions of **M**. Both generative-type models have identical fixed points (Achler & Bettencourt 2011).

In summary, a new method is described where **Y** can be found using **M**. The same solutions to the feedforward equation 1 can be solved by the dynamic generative methods in equations: 4, 5, or 6.

**Recognition Using Expectations**

The next few sections demonstrate that the fixed-point solutions, purpose of networks, and expectations are indicated by the weights in **M**. Since most feedforward networks do not have direct access to the fixed-points, one may assume that determining the fixed points for recognition is difficult. In fact the fixed-points for recognition can be intuitive, easy to define, and learn. The fixed points can be described as the expected input-output behavior of a classifier given well-defined input patterns. To demonstrate this, let's define a matrix that represents expectations and pose a simplified recognition problem.

Suppose we want to discriminate between idealized drawings of a bicycle or unicycle using features of wheels horizontal lines and vertical lines. We can describe the expectation based of these features. The expectation matrix **Exp** is written below.

$$\text{Exp} = \begin{matrix} & X_1 & X_2 & X_3 & X_4... \\ & \begin{bmatrix} 2 & 1 & 1 & 1 \\ 1 & 0 & 0 & 1 \end{bmatrix} & \begin{matrix} Y_1 & \text{Bicycle} \\ Y_2 & \text{Unicycle} \end{matrix} \end{matrix} \quad = M1 \quad (7)$$

Expectation matrix **Exp** indicates characteristic bicycles and unicycles based on features $X_1$= circles (wheels), $X_2$: horizontal lines, $X_3$: handlebar features, $X_4$ : seat features. Although binary values are given they can be any real number. For example if 50% of bicycles have seats then the entry can be 0.5.

Two wheels are expected in a bicycle. Horizontal frame and handlebar features are expected in a bicycle. One wheel is expected and no handlebar features are expected in a unicycle. A matrix such as this is easy to obtain. It only requires the expectation of the features relative to the label to be written. This can be determined by a simple averaging function or co-occurrence of features with labels. It may also be determined by symbolic expressions and language. Although we described a specific example with specific features this can be generalized to any arbitrary features present when a label is present (shown later).

Any supervised classifier trained based on the information above, and given **X**= [1,0,0,1]$^T$, a unicycle, should respond with **Y**=[0,1]$^T$, indicating a unicycle label. **X**= [2,1,1,1]$^T$, bicycle, should generate **Y**=[1,0]$^T$. These values represent the fixed points, ground truth, and critical evaluation points, of the system.

**Optimal Feedforward Weights are Not Representative of Expectation**

Even though supervised weights **W** may store input-label associations, it is not easy to incorporate fixed-point expectations into **W**. To demonstrate, let's assume feedforward weights represent expectations and set **W0 = Exp**. Setting the input to represent a unicycle: $X_{test} = [1,0,0,1]^T$ and solving **Y=W0 $X_{test}$** we get $Y=[3,2]^T$. This is not the expected solution: $Y=[0,1]^T$. **W** should be trained using the expectation matrix as a training set.

### Solving recognition using M

If **M** represents the fixed points one should be able to make **M** equal to the expectation matrix (**M1=Exp**), then insert **M1** and $X_{test}$ into either equations 5 or 6, and obtain the correct solution. When we do this and wait until the dynamics go to zero ($dY/dt \to 0$), the solution obtained for $X_{test} = [1,0,0,1]^T$ is $Y=[0,1]^T$. Unicycle is correctly recognized. The solution for $X_{test}=[2,1,1,1]^T$ is $Y=[1,0]^T$ representing bicycle. Both patterns are correctly recognized demonstrating recognition using the expectation matrix and that the fixed points in this system can be determined by the expectations.

### M vs. W

To demonstrate the relation between **M** and **W**, let's go back to equation 2 and calculate **W1** from **M1** using the pseudoinverse. Since these matrixes may not be square, the standard pseudoinverse method is used where $W=(M^TM)^{-1}M^T$. The transpose **W1** is shown.

$$W1 = \begin{bmatrix} 0.2 & 0.4 & -0.4 & -0.2 \\ 0.2 & -0.6 & -0.6 & 0.8 \end{bmatrix} \begin{array}{l} Y_1 \text{ Bicycle} \\ Y_2 \text{ Unicycle} \end{array} \quad (8)$$

**W** represents the feedforward weights. **W1** is more complex, has negative values, and the values do not clearly indicate fixed-points. To demonstrate that **W1** represents correct feedforward weights, lets calculate **Y=W1 $X_{test}$**. Correct answers are obtained: $Y=[1,0]^T$ for $X_{test}=[2,1,1,1]^T$ and $Y=[0,1]^T$ for $X_{test} = [1,0,0,1]^T$. Correct recognition is obtained either 1) with the feedforward-feedback method (e.g. equation 5) using expectation values from matrix 7, or 2) with the feedforward method (equation 1) using **W** values from matrix 8. The disadvantage of the feedforward method is that **W** is more difficult to obtain and is sub-symbolic.

### Symbolic Information

Suppose we want to ask do bicycles have wheels? How many? These are symbolic questions. Using **M1** we can look up label for bicycle, $Y_1$, and feature for wheel, $X_1$, and read the value: 2. If we want to do the same thing for unicycle we can look up label for unicycle, $Y_2$, and feature for wheel, $X_1$, and read the value: 1. If we attempt this with **W1** we do not retrieve symbolically useful information (0.2 and 0.2). Thus **W** is sub-symbolic but **M** maintains symbolic access while also representing recognition weights. This is possible because **M** represents fixed-point solutions.

### Generalized learning

Before we proceed further, it is important to establish that the feedforward-feedback method can learn equivalent patterns to the feedforward method. This should not be a surprise since they both have the same fixed-points.

In Achler 2012 it is shown that learning of randomly generated data separated by a linear separator can be done and is simpler using **M**. To determine the expectation matrix **M** from the training data, the points corresponding to each label are averaged. The performance was analogous to **W** using a single-layer linear perception. Both methods did not have any testing errors as long as the testing points did not fall on the linear separator. The differences were not in percent correct but in dynamics.

The dynamics consume the most processing time and display variability. The processing time for the dynamic phases of both methods (feedforward during learning and feedforward-feedback during testing) depended on initial conditions, which were randomized. Both require parameters associated with dynamics, for example a criteria to stop iterations. Subsequently given the same data and random initial conditions, the number of iterations required for perceptron learning varied from run-to-run. The number iterations during testing also varied for the feedforward-feedback method. Analogously non-dynamic aspects such as learning times for the feedforward-feedback method and testing times for the perceptron did not vary.

It took the feedforward-feedback method much less time to learn (hundreds of times faster). However, it took longer to perform recognition (about tens of times faster per test). Thus the feedforward-feedback method was faster in learning and the perceptron was faster in testing, but otherwise performance was similar.

The feedforward-feedback architecture has also been used in multiclass classification paradigm using a similar method e.g. (Achler Amir 2008; Achler, Omar, Amir 2008; Achler, Vural, Amir 2009). In that work, the connections were limited to binary values and a single instance was used to train each class. The feedforward-feedback algorithm described here obtains the same results.

**Demonstrating Online Modification**

One of the most important benefits of **M** representing fixed-points is the ease of modification. Suppose after learning **M1** as above, one encounters a new mode of transit, a rollerblade. In the **M** domain a new fixed-point can be added by simply adding another row. This change is localized because nothing else is changed in the rest of the matrix (see **M2** and the new fixed-point in bold). Localized changes are essential for enabling online learning.

$$\mathbf{M2} = \begin{matrix} & X_1 & X_2 & X_3 & X_4 \\ \begin{pmatrix} 2 & 1 & 1 & 1 \\ 1 & 0 & 0 & 1 \\ \mathbf{4} & \mathbf{0} & \mathbf{0} & \mathbf{0} \end{pmatrix} & \begin{matrix} Y_1 & \text{Bicycle} \\ Y_2 & \text{Unicycle} \\ Y_3 & \text{Rollerblade} \end{matrix} \end{matrix} \quad (9)$$

Testing produces the correct results: $\mathbf{Y}=[1,0,0]^T$ for $\mathbf{X}_{test}=[2,1,1,1]^T$, $\mathbf{Y}=[0,1,0]^T$ for $\mathbf{X}_{test}=[1,0,0,1]^T$, and $\mathbf{Y}=[0,0,1]^T$ for $\mathbf{X}_{test}=[4,0,0,0]^T$. If the data was stored in **W**, then what changes would have had to be made using the same example? Lets calculate **W2** using the pseudoinverse:

$$\mathbf{W2} = \begin{bmatrix} 0.0 & 0.5 & 0.5 & 0.0 \\ 0.0 & -0.5 & -0.5 & 1.0 \\ 0.3 & -0.1 & -0.1 & -0.3 \end{bmatrix} \quad (10)$$

Testing produces the correct results as above, however **W2** changed significantly compared to **W1** (see matrix 11 vs. matrix 8). No entry of the original **W1** remained the same. The change of weights is distributed throughout the matrix if recognition information is stored in feedforward weights **W**. The whole network needs to be retrained (using previously learned and new data) to correctly make a distributed change and maintain old fixed-points.

Lets give a more subtle example: suppose one has learned that all bicycles have horizontal features ("men's" bicycle) but now we find that 50% of bicycles do not have a horizontal bar (e.g. "women's" bicycles). This requires modifying the expectation. The new expectation is now:

$$\mathbf{M3} = \begin{matrix} & X_1 & X_2 & X_3 & X_4 \\ \begin{bmatrix} 2 & 0.5 & 1 & 1 \\ 1 & 0 & 0 & 1 \end{bmatrix} & \begin{matrix} Y_1 & \text{Bicycle} \\ Y_2 & \text{Unicycle} \end{matrix} \end{matrix} \quad (11)$$

To reflect this new finding **M1** is changed to **M3** with the highlighted change. Previously entry $M1_{12} = 1$ and now $M3_{12} = 0.5$. This is a relatively minor, localized update. Again lets calculate a new **W3** by taking the pseudoinverse. The solution is:

$$\mathbf{W3} = \begin{bmatrix} 0.3 & 0.3 & 0.6 & -0.3 \\ 0.1 & -0.4 & -0.9 & 0.9 \end{bmatrix} \begin{matrix} Y_1 & \text{Bicycle} \\ Y_2 & \text{Unicycle} \end{matrix} \quad (12)$$

All entries in **W3** are changed from **W1** in order to incorporate the new finding, a distributive change. The whole feedforward network may need retraining for a unitary change in expectation: a single entry. This reveals a fundamental scalability/plausibility problem. If massive feedforward networks are used, for example by the brain for recognition, then all of the connections may have to change for a new piece of information learned or for a change in fixed point. One small modification in expectation may require changes in thousands or millions of entries in **W**. Moreover changes would have to occur immediately in order to instantaneously learn and use that information in an online learning paradigm. Clearly more theoretical work is needed beyond simple examples to determine in what cases this may occur. For example sparse representations may need fewer changes. However the possibility of massive changes is worrisome, especially since changes may occur often. Storing and computing information based on fixed points allows localized modification of **M**, avoiding online learning difficulties associated with distributed modification.

**Cognitive phenomena and Dynamics**

Although equations 1, 4 & 5 have the same solutions and fixed points, the feedforward-feedback equations are dynamic during testing. One can ask how long can recognition take?

For feedforward algorithms there are no dynamics during testing so the answer is simple: the time it takes to recognize is the time it takes to multiply **Y=WX**. Since it is assumed neural components of the brain are parallel the time is constant. This is the basis of the speed-of-sight hypothesis where the speed is the time required for a feedforward pass (Thorpe et al 1996; Serre et al 2007).

The feedforward-feedback dynamics are more complex during testing and so are cognitive findings. The time required for humans appears to follow signal-to-noise ratios (e.g. Duncan & Humphreys 1989; Wolfe 2001; Rosenholtz 2001). When the signal-to-noise ratios are small, then recognition takes longer. These interactions can be measured by the time required to process pattern mixtures. For example, pattern combinations that require different times are shown in figure 1. Signal-to-noise dynamics suggest several cognitive phenomena may be related to feedforward-feedback dynamics. We focus only on a single phenomenon here, difficulty with similarity, to keep this paper brief. A cognitive-focused paper is planned for the future.

**Difficulty with Similarity**

If the patterns are very different, the signal-to-noise ratio is large, and recognition can be faster and independent of the number of patterns present. If the patterns are very similar the signal-to-noise ratio is small and recognition is slower and sensitive to the number of patterns presented. Search difficulty can vary continuously between these extremes based on amount of similarity (Duncan & Humphreys 1989). This phenomenon is fundamental because it can be found in virtually every human recognition modality and task: whether recognizing simple line orientations to complex images.

**Difficulty with Similarity in feedforward-feedback networks**

Signal-to-noise effects and more specifically similarity effects can be demonstrated by numerical simulations and information-theoretic analysis. Lets define a minimal **M** matrix to be composed two fixed points determined by two patterns. Row 1 represents the fixed-point for pattern 1 (just like the bicycle example). If $X_{pattern1}$ is given Y=[1, 0] is expected. Row 2 represents the fixed-point for pattern 2 (like the unicycle example). If $X_{pattern2}$ is given then Y=[0, 1] is expected. We vary the similarity between the patterns and analyze the effects. The pattern data set used in Achler, Vural & Amir (2009) and Achler (2011) is used here because it has a good distribution of variously similar patterns and is motivated by letter patterns. However, randomly generated sets of patterns show the same results. There are 26 patterns and 512 features.

Similarity of a pattern pair is measured by calculating the amplitude of shared feature values, divided by the coverage of all features (plotted on x-axis in fig 2). If the patterns are the same then similarity is 1, if they share no features then similarity is 0. Similarity of pattern pairs in the data ranged between 0.08 to 0.76.

$$\text{Similarity} = \text{sum}(\min(X_{pattern1}, X_{pattern2})) / \text{sum}(\max(X_{pattern1}, X_{pattern2})) \qquad (13)$$

We first show difficulty with similarity using simulations of equations 4 and 5. Then we show that information theoretic methods make the same predictions for these equations.

The number of iterations required in the dynamic equation was obtained by setting $\mathbf{X}_{test}$ to pattern 1 and the initial conditions to $Y_2=1$, $Y_1=0.0001$. The number iterations required until $Y_1=0.9$ was recorded. Then the simulation is repeated but the pairs were switched: now $\mathbf{X}_{test}$ was set to pattern 2 with the initial conditions to $Y_1=1$, $Y_2=0.0001$. The number iterations required until $Y_2=0.9$ was measured. The number of iterations from both conditions were averaged together. The number of iterations of *least squares* equation 4, and *regulatory feedback* equation 5, are plotted relative to similarity (x-axis) in figure 2. $dt=0.02$ was used for the least squares algorithm to insure convergence. In least squares algorithm this parameter may need to be changed with the number of features in $\mathbf{X}$. The stability the regulatory feedback algorithm is not sensitive to *dt* as long as *dt*<=1. Further comparisons between different feedforward-feedback algorithms are the focus of future work.

Beyond the simulations, we can show that difficulty with similarity trends follow information theoretic predictions. The feedforward feedback method can be thought of as performing several steps at once during testing: using $\mathbf{M}$ to estimate $\mathbf{W}$ for the given test patterns and performing recognition. This is why we can also use information-theoretic methods that estimate difficulty of taking the pseudoinverse of matrix, namely the condition number κ, to estimate the difficulty of discrimination. The condition number κ is defined as:

$$\kappa = ||\mathbf{M}^{-1}|| \cdot ||\mathbf{M}|| \quad \text{or} \quad \kappa = ||\mathbf{W}|| \cdot ||\mathbf{M}|| \quad \text{or} \quad \kappa = ||\mathbf{W}^{-1}|| \cdot ||\mathbf{W}|| \quad (14)$$

A large number predicts a more-difficult pseudoinverse. As the rows in $\mathbf{M}$ are more similar to each other, the condition number increases. The condition numbers for the data set varied between 1.17 (left) and 3.68 (right), and increased with similarity (fig 2 green trace). The similarity of pattern 1 and pattern 2 pairs, condition number of $\mathbf{M}$ composed of the pairs, and the number of iterations required for recognition are plotted in figure 2.

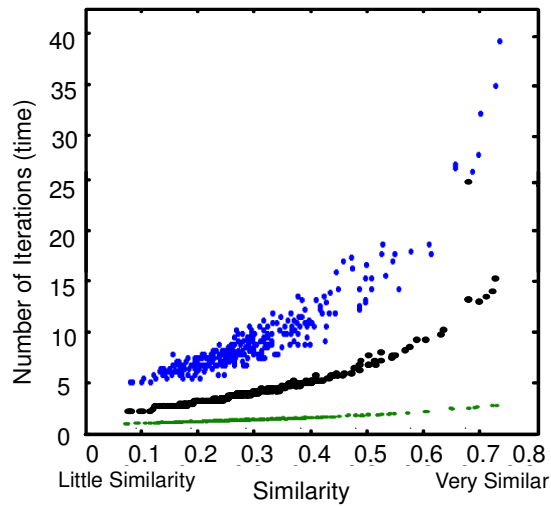

**Figure 2: Difficulty of Pattern Pairs Based on Similarity.** Condition numbers (*bottom,green*) and the number of iterations in Least Squares (*top,blue)* and Regulatory Feedback (*middle,black*), are plotted relative to similarity. If patterns are similar (right) the number of iterations and condition number increases.

The graph shows that as the patterns are more similar, the condition number and number of iterations increase in both versions of feedforward-feedback algorithms. Thus the condition number of $\mathbf{M}$ and both testing based recognition models are sensitive to similarity affects.

The estimated time it takes to spike and transfer information is 1-3 ms per neuron for neurons localized in the brain. In the feedforward-feedback circuit there are two neurons: a feedforward and feedback neuron. This means that each iteration could take between 2-6ms. Thus the amount of time to recognize is estimated about 12-150 ms. This is a rough estimate because it does not include the time it may take to speak or move a button, the time for spatial search when search becomes difficult, or the time to transfer information from the eyes to the brain. Since responses can range between about 170ms to seconds these findings are within the neutrally plausible range. However the most significant implication is that in the feedforward-feedback recognition time is inherently associated with similarity effects, mimicking cognitive phenomena, while in the feedforward it is not.

**Discussion**

For equations 1-6, **M** describes the fixed points or solutions of the network. There are several advantages of using connection weights **M**. 1) For learning, determining expectation is simpler than implementing a delta-rule. 2) Learning a new fixed point or expectation can be as simple as modifying or adding a row to **M** without changing other patterns. This enables online learning. 3) The fixed points allow symbolic access to the underlying associations of the network. 4) The dynamics during recognition emulate cognitive phenomena.

The most important finding presented here is that the form of information stored in **M** represents fixed points. Since **M** represents the fixed points, then by definition any supervised modification of a fixed point – e.g. incorporating a new piece of labeled information obtained during an online learning instance –is localized. This means that fixed-points that were not modified remain unaltered and the network does not need batch-learning to rehearse old fixed points.

The fixed-points form a bridge between weights favorable for recognition and weights favorable for symbolic representations. With fixed point weights it is possible to manipulate the representations and perform symbolic analysis on the stored patterns. Moreover knowledge can be directly ascertained from the weight matrix, for example answering do bicycles have wheels. Like adding a new fixed point the localized property can allow parts of matrixes to be recombined and reused with other matrixes to form new matrixes. Such functions are referred to as recognition logic and are to be elaborated further in future work.

Fixed points are intuitive and easy to learn. **M** can be determined by the expectation of the features relative to the label. This does not require feedback dynamics or error propagation during learning. Expectation can be determined by a simple averaging function, averaging the number of times a feature is present when a label is present, as was done in the learning example presented here. Other possible methods may calculate label-to-feature correlation, co-occurrence, or calculate the likelihood a feature is present when a label is present. The fixed points may even be determined based on symbolic-like descriptions of features-to-labels relationships. Other implementations may use a simple Hebbian-type learning, hierarchy, focus on reducing condition numbers, and/or avoid linearly dependent situations. These methods are to be further explored in future work.

The feedforward-feedback model was originally designed as a method to simplify and perform recognition with symbolic like weights. Subsequently it was found to improve online learning and scalability. Analyzing its limitation due to dynamics it is found to *inherently* display certain cognitive phenomena associated with recognition. The cognitive phenomena are inherent because no design or parameters went into displaying these properties. Feedforward classifiers do not *inherently* display similarity properties however there is a rich literature modeling similarity e.g. (Duncan & Humphreys 1989; Bundesen 1990; Usher & Niebur 1996; Breitmeyer & Ögmen 2006; Serre et al 2007; Hermens et al. 2008; Ratcliff & McKoon 2008). Such models have dedicated parameters for competition, hierarchy, relatedness, processing time, modality, or visual space. Instead of designing a model to account for increased time with similarity, such properties are shown as fundamental to the feedforward-feedback algorithm that facilitates symbolic processing and online learning. Thus the feedforward-feedback model presents a more integrated, general, and unified account.

It is important to realize that from a computational perspective, costs and benefits are associated with both the feedforward and reconstruction models. It can be argued that the feedforward model is more efficient during the testing phase since it requires one pass of information per test. In contrast, the feedforward-feedback recognition is more-complex, dynamic, and requires at least three passes of information processing: feedforward, feedback and feedforward. On the other hand, feedforward models are not as efficient in learning: requires multiple passes, training is more-complex, training independent new fixed points without retraining ultimately results in catastrophic interference and forgetting, the fixed points are opaque and sub-symbolic.

Since dynamics are involved in feedforward learning of **W**, it is fair to ask the analogous question of how long can learning take? Analogous to the feedforward-feedback testing, certain pattern combinations in feedforward learning should take longer to learn. Subsequently, learning these combinations may take longer to converge, but with batch learning all of the patterns are learned at once. The larger the data sets the more likely learning difficulties are encountered in batch mode, slowing large network learning (Bottou & Le Cun 2004). In the feedforward-feedback method learning is online and pattern combinations are more independent during testing. Even if two patterns combinations are indistinguishable and slower to converge during testing, this is not as disruptive because in those cases these nodes interact with the rest of the network as one node. Even if those patterns are slower to converge the others remain fast (Achler 2011).

In light of these limitations, if recognition is simple and performed in a constrained environment, then a feedforward model may be more beneficial. On the other hand, if recognition requires flexibility, and successive changes throughout learning, then a reconstruction model may be more beneficial. It may be possible that the brain may start with a reconstruction model and use it to train a feedforward model as a task becomes automatic and learning becomes less variable.

There is still much work to be able to perform recognition like a human. Current methods suffer from difficulties of recognition in the real world involving: shading, lighting, size or rotation invariance, size invariance, and so on. However, this paper suggests a solution for a small part of the puzzle outlining how brain-like connectionist networks may make only localized online changes, quickly learn new information, and have symbolic access to learned information.